\documentclass[sigconf]{acmart}

\usepackage{booktabs}
\usepackage{subcaption}
\usepackage{enumitem}
\usepackage{pifont}
\usepackage{tikz}
\usepackage{float}
\usetikzlibrary{positioning, arrows.meta, fit, backgrounds, calc}

\newcommand{\cotomiact}{cotomi Act}
\newcommand{\captionarchitecture}{Unlike conventional computer-using agents that execute tasks with no knowledge of the user's organization, \cotomiact{} continuously learns from ordinary browsing. The Behavior Logger (top) monitors user activity and distills it via an agentic ETL pipeline into structured artifacts---tasks, timelines, and wiki pages---stored in a shared Knowledge Workspace (bottom) that both the user and the agent can read and edit. The Agent (right) consults these artifacts during its ReAct loop, acting as a situated co-worker rather than a stateless executor.}

\copyrightyear{2026}
\acmYear{2026}
\setcopyright{cc}
\setcctype{by}
\acmConference[CAIS '26]{ACM Conference on AI and Agentic Systems}{May 26--29, 2026}{San Jose, CA, USA}
\acmBooktitle{ACM Conference on AI and Agentic Systems (CAIS '26), May 26--29, 2026, San Jose, CA, USA}
\acmDOI{10.1145/3786335.3813203}
\acmISBN{979-8-4007-2415-2/2026/05}
\title{cotomi Act: Learning to Automate Work by Watching You}

\makeatletter
\newcommand{\sharedcontactblock}{%
  \vspace{10pt}%
  {\centering\@authorfont NEC Corporation\par}%
  \vspace{2pt}%
  {\centering\small\ttfamily
    \{oyamada, k\_takeoka, kosuke\_a, ryoma-obara,\par
    masafumi-enomoto, haochen-zhang, daichi-haraguchi, tamura-takuya\}@nec.com\par}}
\newcommand{\authorbreak}{\g@addto@macro\addresses{\authorbreakmarker}}
\newcommand{\authorbreakmarker}{\global\let\and\@typeset@author@bx}
\patchcmd{\@mkauthors@iii}
  {\addresses\let\and\@typeset@author@bx\and\par\bigskip}
  {\addresses\let\and\@typeset@author@bx\and\par\sharedcontactblock\bigskip}
  {}{}
\patchcmd{\@mkauthors@iv}
  {\addresses\let\and\@typeset@author@bx\and\par\bigskip\egroup}
  {\addresses\let\and\@typeset@author@bx\and\par\sharedcontactblock\bigskip\egroup}
  {}{}
\makeatother

\author{Masafumi Oyamada}\authorbreak
\author{Kunihiro Takeoka}\authorbreak
\author{Kosuke Akimoto}\authorbreak
\author{Ryoma Obara}\authorbreak
\author{Masafumi Enomoto}\authorbreak
\author{Haochen Zhang}\authorbreak
\author{Daichi Haraguchi}\authorbreak
\author{Takuya Tamura}

\begin{abstract}
What if a browser agent could learn your work simply by watching you do it?
We present \cotomiact{}, a browser-based computer-using agent that combines reliable multi-step task execution with persistent organizational knowledge learned from user behavior.
For execution, an agent scaffold with adaptive lazy observation, verbal-diff-based history compression, coarse-grained actions, and test-time scaling via best-of-N action selection achieves 80.4\% on the 179-task WebArena human-evaluation subset, exceeding the reported 78.2\% human baseline.
For organizational knowledge, a behavior-to-knowledge pipeline passively observes the user's browsing and progressively abstracts it into artifacts (task boards, wiki) exposed through a shared workspace editable by both user and agent.
A controlled proxy evaluation confirms that task success improves as behavior-derived knowledge accumulates.
In our live demonstration, attendees interact with the system in a real browser, issuing tasks and observing end-to-end autonomous execution and shared knowledge management.
\end{abstract}



\begin{document}

\renewcommand{\shortauthors}{Oyamada et al.}
\maketitle

\begin{figure}[t]
  \centering
  \includegraphics[width=\columnwidth]{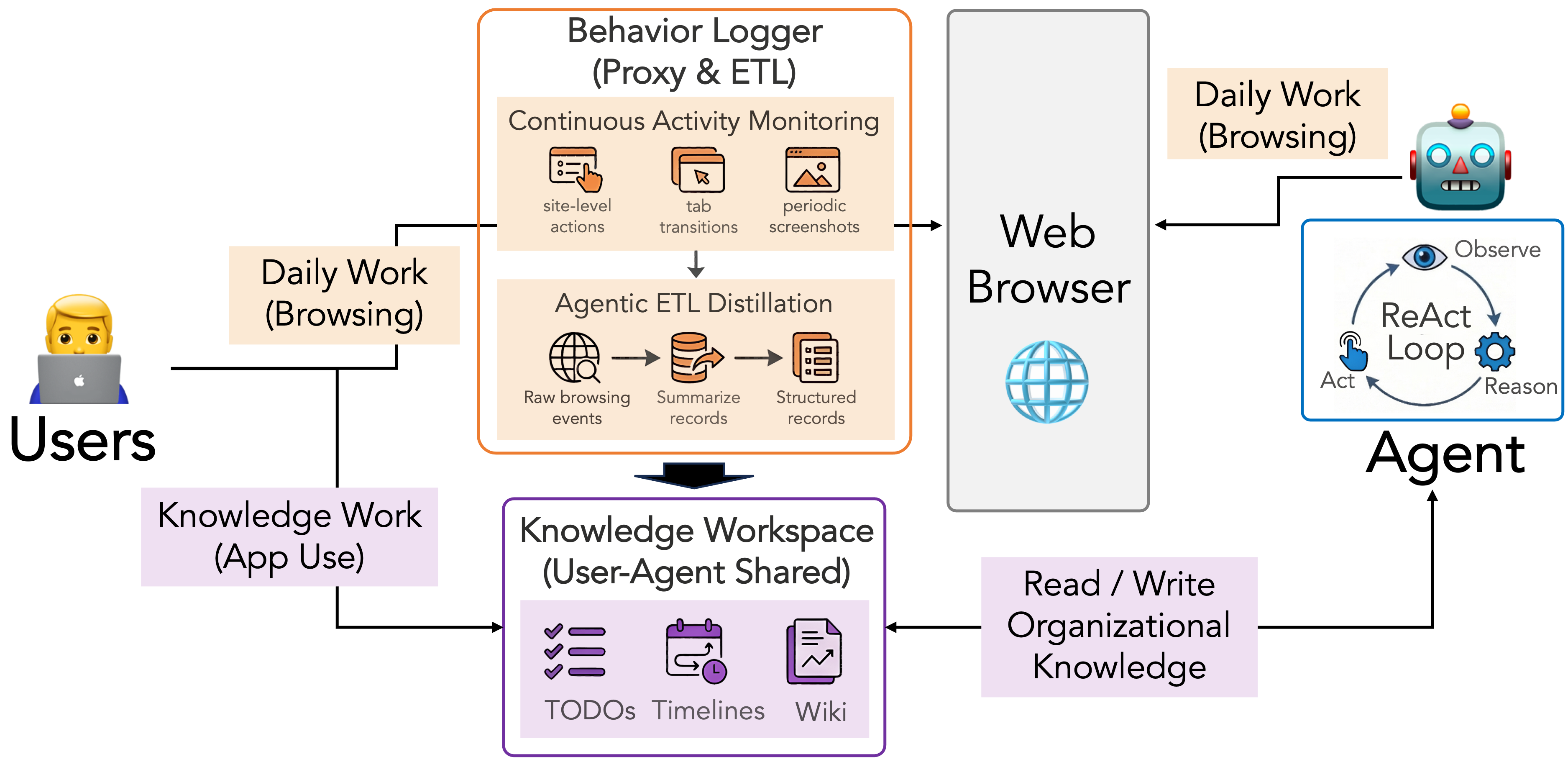}
  \caption{\captionarchitecture}
  \label{fig:architecture}
\end{figure}

\section{Introduction}
\label{sec:introduction}

Large language models (LLMs) have grown dramatically more capable at reasoning, planning, and tool use, giving rise to \emph{computer-using agents} (CUAs) that can operate web browsers on behalf of users~\cite{openai2025cua,anthropic2024computeruse,google2024mariner}.
Yet strong reasoning alone is insufficient for a reliable co-worker.
Every organization operates with tacit knowledge---who approves what, what internal jargon means, and which tasks are already in progress---that is rarely written down but essential for getting things done.
Today's computer-using agents ignore this reality.
To function as a genuine co-worker, a CUA needs two ingredients that today's systems still struggle to combine:
\textbf{(1)~strong reasoning ability} for reliably carrying out complex, multi-step web tasks, and
\textbf{(2)~organizational knowledge}---the tacit preferences, unwritten workflows, and conventions that exist in every organization but are rarely documented.

Prior systems fall short on both counts.
On the \textbf{reasoning} side, even well-specified benchmark tasks remain difficult: the best reported agent reaches 71.6\% on WebArena~\cite{guo2026opagent}, still below the 78.2\% human baseline~\cite{zhou2024webarena}.
Long-horizon tasks in complex web UIs demand repeated observation and action, producing lengthy context that strains model reasoning.
On the \textbf{organizational-knowledge} side, current CUAs execute each task knowing nothing about local workflows, preferences, or pending work.
Without such context, an agent faces a costly dilemma on every ambiguous instruction: ask frequent clarification questions at the expense of efficiency, or silently guess and risk irreversible errors.
Moreover, organizational knowledge spans a wide spectrum---from relatively stable conventions and approval rules to rapidly changing states such as who is assigned to which task right now.
A natural response is to let the agent retrieve internal documents via RAG-like mechanisms~\cite{xu2025raggui}, but enterprise repositories are vast and often stale, and much of the knowledge that matters most---how people actually work---is never written down at all.

We present \cotomiact{}, a browser-based CUA that addresses both requirements.
For \textbf{strong reasoning ability}, the system employs a carefully engineered agent scaffold (Section~\ref{sec:architecture}) combining adaptive lazy observation, verbal-diff-based history compression, coarse-grained actions, and test-time scaling via best-of-N action selection and task decomposition~\cite{komiyama2026bestofn,light2025disc}---together keeping the context window focused and enabling reliable execution over long-horizon web tasks.
On WebArena~\cite{zhou2024webarena}, this scaffold achieves 80.4\% on the 179-task human-evaluation subset~\cite{nec2025cotomiact}, exceeding the reported human baseline of 78.2\%~\cite{zhou2024webarena} (Table~\ref{tab:webarena}).
For \textbf{tacit / organizational knowledge}, \cotomiact{} takes a different approach: it \emph{learns from user behavior}. A background activity monitor observes the user's ordinary browsing, and an LLM-based pipeline (Section~\ref{sec:architecture}) progressively distills those observations---which documents the user consults, in what context, and what tasks they are working on---into a working model of how the user works. This lets the agent navigate the vast sea of enterprise information along paths the user has already carved. The resulting knowledge is surfaced as shared artifacts such as task boards and wiki pages that both the user and the agent can read and edit; through routine use, the user naturally adds, corrects, and removes entries, so that even undocumented knowledge finds its way into the agent's grounding.

Our contributions are:
\begin{enumerate}[nosep,leftmargin=*]
  \item \cotomiact{}, a browser-based computer-using agent that achieves 80.4\% on the 179-task WebArena human-evaluation subset---exceeding the reported human baseline (78.2\%)---through an agent scaffold combining adaptive lazy observation, verbal-diff-based history compression, coarse-grained actions, and test-time scaling via best-of-N action selection over multi-step web tasks.
  \item A behavior-to-knowledge pipeline that passively observes a user's ordinary browsing, progressively abstracts raw events into reusable knowledge artifacts (task boards, wiki, activity timelines), and exposes them through a shared workspace where both the user and the agent can read and edit---enabling persistent learning and transparent human--agent alignment.
  \item A controlled proxy evaluation showing that downstream task success generally improves as the agent accumulates more behavior-derived knowledge---empirically validating the behavior-to-knowledge pipeline.
\end{enumerate}
The remainder of this paper describes the system architecture (Section~\ref{sec:architecture}), evaluation (Section~\ref{sec:evaluation}), and live demonstration scenario (Section~\ref{sec:demo}).

\section{System Architecture}
\label{sec:architecture}

Figure~\ref{fig:architecture} shows the architecture of \cotomiact{}.
The design mirrors the two requirements identified in Section~\ref{sec:introduction}: an \emph{agent scaffold} that provides strong reasoning over multi-step web tasks, and a \emph{behavior-to-knowledge pipeline} that continuously converts observed user activity into organizational knowledge the agent can act on.
A \emph{shared knowledge workspace} bridges the two, storing artifacts that both the user and the agent read and edit.
All three are delivered through a single browser extension, with the agent scaffold and behavior logger running as backend services described below.

\subsection{Agent Scaffold}
The web automation agent follows a ReAct-style loop~\cite{yao2023react} of observation, reasoning, and action, orchestrated by what we call the \emph{agent scaffold}---the infrastructure that manages browser state, composes prompts, and routes actions.
At each step, the model's context window carries two kinds of information: the \textbf{current observation} (what the agent sees right now) and the \textbf{execution history} (what happened in prior steps).
Both compete for the same finite token budget, and their design interacts with the \textbf{action space} that determines how many steps a task requires.
The central challenge is \emph{context bloat}: feeding the model a full accessibility tree at every step, or retaining all past observations verbatim, fills the context window with stale information and degrades decision quality in later steps.

\begin{figure}[t]
  \centering
  \includegraphics[width=0.95\columnwidth]{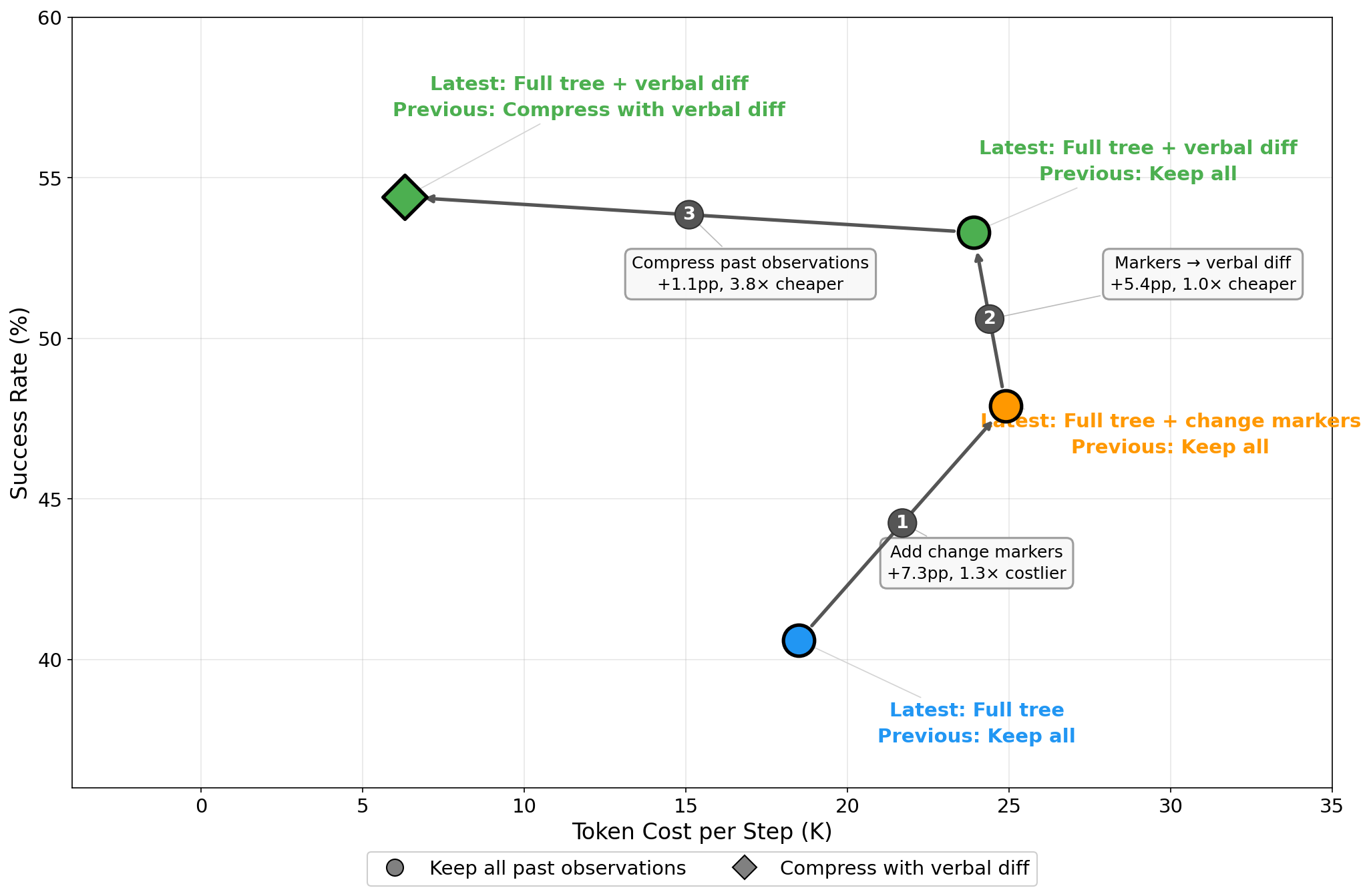}
  \caption{Observation and history design trajectory on WebArena-Verified (Gemma-4-31B-IT, 82 tasks, 3 runs). Each arrow traces one design change in accuracy--token-cost space. Applying verbal diffs to the current observation improves accuracy, while using them to replace stale history reduces cost, jointly moving the scaffold toward the upper-left (higher accuracy, lower cost).}
  \label{fig:trajectory}
\end{figure}

We address this through four scaffold-level mechanisms: adaptive observation, compact execution history, coarse-grained actions, and task decomposition.
We validate the observation and history choices through a systematic ablation on WebArena-Verified~\cite{hattami2025webarenaverified} (Gemma-4-31B-IT, 39~configurations, 82 tasks $\times$ 3 runs; Figure~\ref{fig:trajectory}).
We then describe how the scaffold retrieves organizational context from the shared workspace.

\paragraph{Current observation.}
A richer current observation generally improves accuracy.
Starting from a baseline of the full accessibility tree with no change information (40.6\% SR), adding rule-generated \emph{verbal diffs}---natural-language descriptions of
DOM-level changes (e.g., ``New element: [42] button `Submit'\,'')---yields
the largest single improvement (+12.7\,pp to 53.3\%), substantially
outperforming positional change markers (+7.3\,pp).
However, richer observations increase per-step token cost and latency.
When latency matters more than peak accuracy, a \textbf{lazy observation} strategy offers a compelling trade-off: sending only the viewport-visible elements---and letting the agent request the full tree on demand---achieves comparable accuracy while reducing end-to-end response time
from 471\,s to 184\,s per task (roughly 2.6$\times$ faster).
\cotomiact{} adopts this lazy strategy by default and enriches on demand for visually complex pages.

\paragraph{Execution history.}
Retaining full past observations in the conversation is counterproductive: input tokens grow linearly with steps and reach ${\sim}$57K by step~20, crowding the context window.
The key insight is that the agent does not need the raw page state at step~5---it needs to know \emph{what changed} between steps~4 and~5.
Replacing stale observations with the same verbal diffs used in the current observation achieves comparable accuracy (+1.1\,pp) while reducing per-step token cost by 3.8$\times$ (23.9K $\to$ 6.3K).
While diff-based history representations have been shown to be token-efficient~\cite{enomoto2026readmorethinkmore}, we find that these natural-language change descriptions outperform standard diff formats by preserving semantic context rather than positional deltas.
Notably, the current-observation and history axes partially substitute: both encode ``what changed'' through different channels, so their combined gain (+13.8\,pp) is less than the sum of individual effects (+22.1\,pp).

\paragraph{Coarse-grained actions.}
If the action space is too fine-grained, the agent needs many steps to accomplish a single intent; the resulting long context dilutes relevant information and lowers task success.
If actions are too coarse, the agent loses the flexibility to handle unexpected page layouts.
We resolve this by defining actions at the highest abstraction that still covers common web interactions.
Scrolling illustrates the point: replacing pixel-level \verb|scrollBy(x, y)|---which can require tens of repetitions to reach a target---with element-level \verb|scrollInto(element)| reduces scroll sequences to a single step without sacrificing precision.

\paragraph{Task decomposition.}
Long-horizon tasks risk \emph{context overflow}: as steps accumulate, earlier observations crowd the reasoning window and degrade decision quality.
To counteract this, the scaffold decomposes a task when the reasoning model emits a structured plan indicating multiple independent sub-goals~\cite{light2025disc}, spawning parallel sub-agents that each maintain a focused, compact context (see Appendix~\ref{app:scaffold} for details).

\paragraph{Organizational context.}
Beyond controlling what the model sees during execution, the scaffold also controls what the agent knows \emph{before} it acts.
When organizational context is needed, the reasoning model explicitly invokes a \texttt{search\_workspace} tool that accepts a natural-language query and returns matching task items, wiki pages, or activity-timeline entries ranked by relevance.
The model decides autonomously whether and when to invoke this tool---typically when the user instruction is under-specified, references organizational jargon, or asks the agent to resume pending work.
This on-demand design avoids injecting irrelevant context on every turn and bridges the gap between a generic executor and a situated co-worker.
The next subsection describes how these workspace artifacts are populated
from observed user behavior.

\subsection{User Behavior Logger}
Learning how the user works requires observing their actual behavior, but raw browsing logs are noisy, voluminous, and far too low-level to be useful as organizational knowledge.
The behavior logger addresses this through a two-phase design: lightweight capture followed by progressive LLM-based abstraction.

In the capture phase, the activity monitor records site-level actions, tab transitions, and periodic screenshots in the background, with minimal interference to the user's workflow.
A \emph{diff-based compression} scheme deduplicates consecutive observations on the same site, reducing both storage volume and downstream noise.

In the abstraction phase, an LLM-based ETL pipeline periodically processes compressed events through three stages:
\begin{enumerate}[nosep,leftmargin=*]
  \item Segment raw events into task-level episodes using inactivity timeouts.
  \item Summarize each segment into a structured record (screen state, actions, resulting changes) and classify it by activity category.
  \item Aggregate summaries into knowledge artifacts---task items, wiki pages, and activity timelines.
\end{enumerate}
We compared a fixed-rule pipeline against an \emph{agentic ETL} design in which the LLM autonomously decides segmentation boundaries, abstraction level, and artifact routing.
In a blind comparison by two annotators across 50~browsing sessions, the agentic variant produced more accurate and coherent artifacts, an important advantage given the volume and heterogeneity of real browsing logs.

\paragraph{Privacy.}
The behavior logger operates on a strictly \emph{opt-in} basis: recording is disabled by default and activates only after explicit user consent, and the user can pause or stop recording at any time.
All captured data is processed through a PII-masking stage that redacts personally identifiable information before artifacts are written to the shared workspace.

\begin{figure}[t]
  \centering
  \includegraphics[width=\columnwidth]{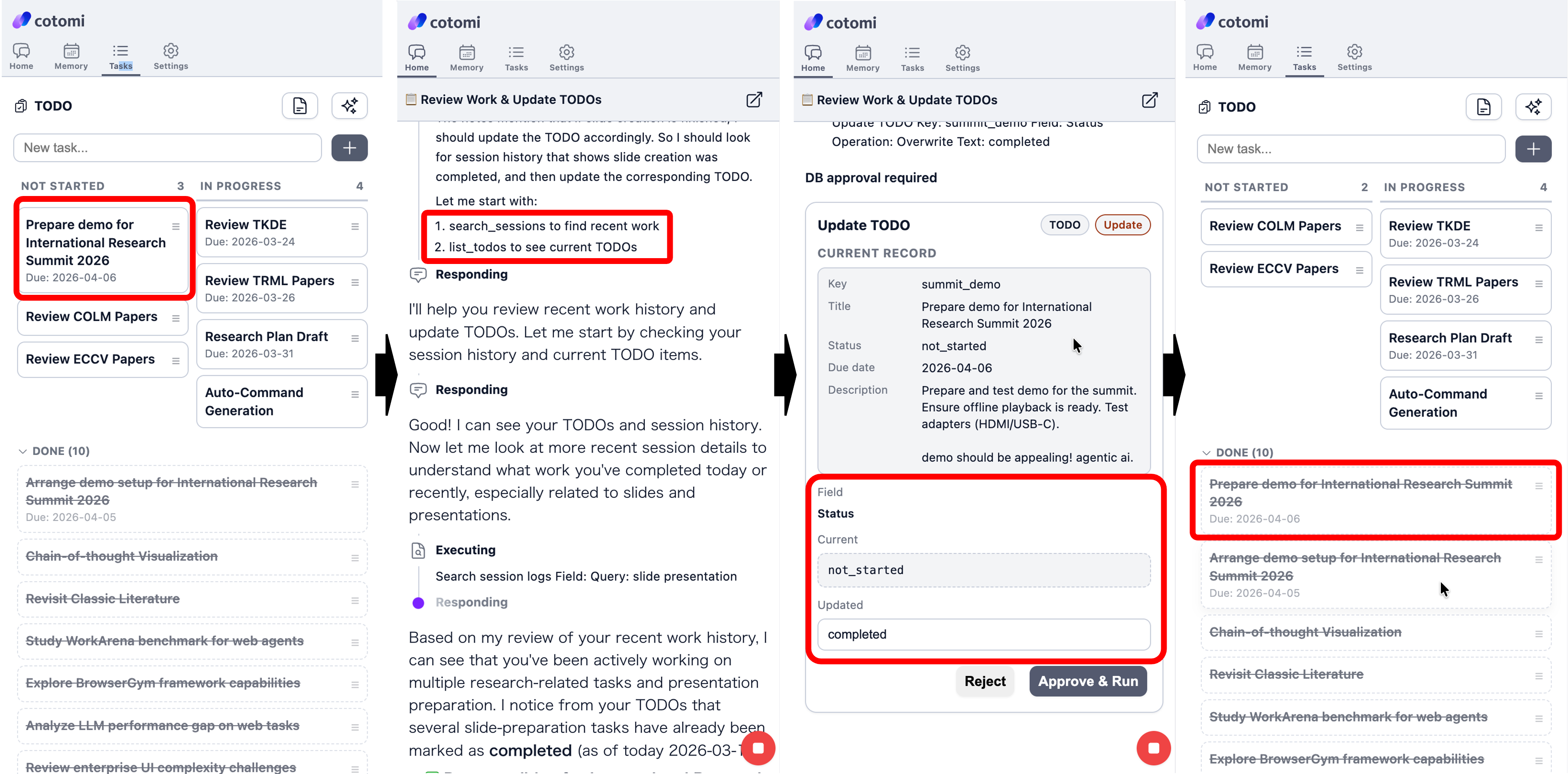}
  \caption{Bidirectional workspace curation in action. The agent reviews the user's recent browsing sessions, proposes a task status update (not\_started $\to$ completed), and the user approves the change---keeping the shared workspace aligned with reality.}
  \label{fig:etl-todos}
\end{figure}

\subsection{Shared Knowledge Workspace}
The central design principle of the workspace is that every knowledge artifact is a \emph{first-class user application}---simultaneously human-readable and agent-queryable---rather than an agent-internal memory~\cite{packer2023memgpt,wang2023voyager,wang2024workflowmemory}.
This makes bidirectional editing natural: the agent bootstraps artifacts from observed behavior, and the user inspects, edits, and approves them through familiar interfaces.
The workspace currently provides three artifact types:
\begin{itemize}[nosep,leftmargin=*]
  \item \textbf{Activity timeline}: a calendar-based display annotated with activity tags (e.g., ``communication,'' ``research,'' ``reporting'') and durations; users can drill down into any tag to review detailed descriptions.
  \item \textbf{Task board}: a shared task database that both agent and user can read and write---the agent proposes updates by reviewing recent activity, while the user retains approval authority over every change.
  \item \textbf{Wiki}: a collection of pages capturing organizational rules, conventions, and procedural notes (e.g., approval workflows, naming conventions, preferred vendors) that the agent distills from observed behavior and the user curates over time.
\end{itemize}

This bidirectional loop (Figure~\ref{fig:etl-todos}) is essential to organizational understanding: the agent's inferences from observation alone are inevitably incomplete, so exposing artifacts for human curation keeps the agent's knowledge aligned with the user's evolving reality.

\section{Evaluation}
\label{sec:evaluation}

We evaluate \cotomiact{} against the two ingredients introduced in Section~\ref{sec:introduction}: reliable task execution (Section~\ref{sec:eval:webarena}) and behavioral knowledge (Section~\ref{sec:eval:behavior}).

\subsection{Task Execution}
\label{sec:eval:webarena}

\begin{table}[t]
  \centering
  \caption{Task success rate (\%) on WebArena.
  Human performance was measured on a 179-task subset of the 812-task benchmark~\cite{zhou2024webarena}; we evaluate on the same subset (\emph{Human subset}) to enable direct comparison, and additionally report full-benchmark scores (\emph{All tasks}).
  $\dagger$Recomputed by us from publicly available results.
  ``site'' = site-specific prompts covering UI elements and navigation; ``fmt'' = answer-format clarifications only.}
  \label{tab:webarena}
  \small
  \begin{tabular}{lccc}
    \toprule
    \textbf{System} & \textbf{Hints} & \textbf{All tasks} & \textbf{Human subset} \\
    \midrule
    SteP~\cite{sodhi2024step} & site & 33.5 & 40.8$^\dagger$ \\
    OpenAI Operator~\cite{openai2025cua} & site & 58.1 & --- \\
    CUGA~\cite{shlomov2026cuga} & site & 61.7 & 64.3$^\dagger$ \\
    OpAgent~\cite{guo2026opagent} & site & 71.6 & 74.9$^\dagger$ \\
    \midrule
    Human~\cite{zhou2024webarena} & --- & --- & 78.2 \\
    \midrule
    \cotomiact{} (base) & fmt & 74.4 & 76.5 \\
    \textbf{\cotomiact{} (+TTS)} & fmt & 75.7 & \textbf{80.4} \\
    \bottomrule
  \end{tabular}
\end{table}

We evaluate the execution engine on WebArena~\cite{zhou2024webarena}, a benchmark of 812 realistic web tasks spanning five domains (e-commerce, forums, content management, maps, and GitLab).

\smallskip\noindent\textbf{Setup.}
WebArena tasks are well-specified single-session instructions that isolate execution capability from organizational context, making it a suitable testbed for the execution requirement.
To ensure a fair comparison with the human baseline, we evaluate on the same 179-task subset used for the human performance study~\cite{zhou2024webarena}.\footnote{Human trajectories: \url{https://github.com/web-arena-x/webarena/blob/main/resources/README.md}}
Because we found that the WebArena automatic scorer produces false positives and false negatives on a non-trivial fraction of tasks, three annotators manually verified every automatic judgment~\cite{zhang2025webarenamod}.
Table~\ref{tab:webarena} compares \cotomiact{} against representative systems.
Notably, \cotomiact{} (marked ``fmt'') uses only answer-format clarifications, without any site-specific UI or navigation hints used by other systems.

\smallskip\noindent\textbf{Results.}
\cotomiact{} achieves a success rate of 80.4\% on this 179-task subset (75.7\% on the full benchmark), exceeding the reported human baseline of 78.2\%~\cite{zhou2024webarena}.
The base scaffold---adaptive lazy observation, verbal-diff-based history compression, and coarse-grained actions---already reaches 76.5\%.
Adding test-time scaling (+TTS), i.e.\ best-of-N action selection and task decomposition~\cite{komiyama2026bestofn,light2025disc}, yields a further 3.9 percentage-point gain, confirming that reducing per-step variance through pre-execution consensus is critical for long-horizon web tasks (Appendix~\ref{app:scaffold} provides additional details).

\subsection{Behavioral Knowledge}
\label{sec:eval:behavior}

We evaluate whether accumulated behavioral knowledge improves task success, and which format of knowledge is most effective, using a controlled study on WorkArena-L1~\cite{drouin2024workarena}.

\smallskip\noindent\textbf{Setup.}
WorkArena-L1 tasks resemble repetitive knowledge-work procedures in enterprise systems (filtering records, filling forms, ordering catalog items), making them sensitive to procedural familiarity rather than exploratory reasoning.
We collect successful agent trajectories across WorkArena-L1's six task categories as a proxy for observed user behavior.
This is a controlled approximation: real user traces would additionally contain failures, interruptions, and exploratory detours, so the pipeline's robustness to such noise remains to be validated in a full deployment study.
We vary \emph{domain coverage}---the fraction of categories whose behavioral knowledge is available to the agent---from 0\% (no knowledge) to 100\% (all six categories, ${\sim}277$ hints total); see Appendix~\ref{app:protocol} for full protocol details.
The three formats---\emph{trajectory} (full trace), \emph{script} (procedural steps), and \emph{insight} (abstract summary)---represent progressive abstraction levels produced by the agentic ETL pipeline (Section~\ref{sec:architecture}); all are stored as wiki pages and retrieved on demand via the scaffold's workspace-search tool.

\begin{figure}[t]
  \centering
  \includegraphics[width=0.80\columnwidth]{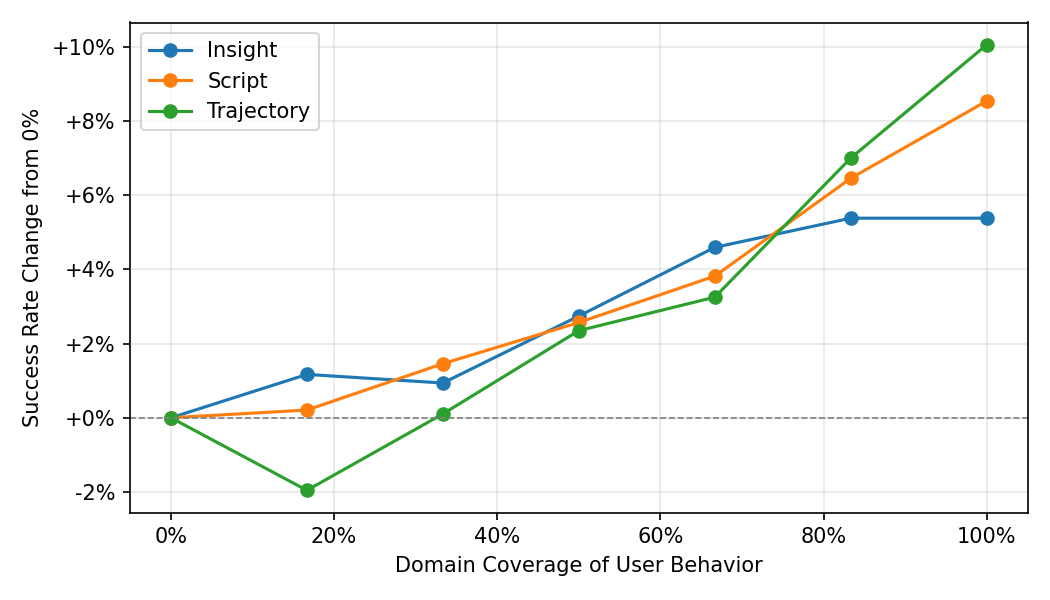}
  \caption{Change in success rate relative to the zero-coverage baseline (i.e., the agent uses no behavioral knowledge) as domain coverage of user behavior increases. More coverage improves downstream task success, while the most effective format depends on how much user behavior is available.}
  \label{fig:behavior-coverage}
\end{figure}

\smallskip\noindent\textbf{Results.}
Figure~\ref{fig:behavior-coverage} shows that task success generally improves with coverage for all three formats, gaining up to +10 percentage points over the zero-coverage baseline of ${\approx}51\%$ (each point averaged over six random category orderings; see Appendix~\ref{app:protocol} for protocol details)---confirming that behavioral knowledge is broadly beneficial.
The most effective format, however, depends on how much user behavior is available: raw trajectories actually \emph{degrade} performance at low coverage (--2\,pp at 17\% coverage) before recovering to become the strongest format at high coverage, whereas scripts and insights remain stable across the entire range.

\smallskip\noindent\textbf{Implications for system design.}
Raw trajectories achieve the highest ceiling but are impractical at scale; the abstraction--detail trade-off motivates \cotomiact{}'s agentic ETL pipeline, which distills raw traces into compact, retrievable artifacts.

A concrete episode illustrates the full observe--extract--act chain.
In the service-catalog domain, the behavior logger records a user completing catalog-ordering tasks (navigating to \emph{Service Catalog $\to$ Hardware}, selecting a product, filling configuration fields, and clicking \emph{Order Now}).
The agentic ETL pipeline segments these sessions and distills each into a procedural script; 48 such scripts are generated across all catalog-ordering templates.
When the agent subsequently encounters a catalog-ordering task, it invokes \texttt{search\_workspace}, retrieves the matching script, and follows its steps.
Before receiving any catalog-ordering scripts the agent succeeds on ${\approx}75\%$ of these tasks; after the 48~scripts become available, success jumps to ${\approx}99\%$ (averaged across six random category orderings).
This sharp improvement confirms that the behavioral-knowledge pipeline produces artifacts with direct causal impact on task execution.
Because these artifacts live in the shared knowledge workspace, the user can inspect and correct them, closing the gap that pure abstraction inevitably opens.

\section{Demonstration Scenario}
\label{sec:demo}

The live demonstration allows attendees to interact with \cotomiact{} through a browser equipped with the extension, connected to a hosted agent backend.
Three scenarios showcase the system's core capabilities in approximately 10 minutes.

\smallskip\noindent\textbf{Scenario 1: Ad-hoc task automation.}
Attendees type a natural-language instruction (e.g., ``Search for recent AI conferences in 2026 and list the submission deadlines'') and observe the agent autonomously navigating, extracting information, and recovering from errors in real time.

\smallskip\noindent\textbf{Scenario 2: Inspecting the shared knowledge workspace.}
Using a pre-populated activity history, the demonstrator shows the calendar-based memory view with time-use analytics, drills down into a specific activity tag, and inspects generated artifacts such as task items and wiki entries.

\smallskip\noindent\textbf{Scenario 3: Workspace-guided execution.}
The demonstrator edits a task item to refine its priority or add contextual notes, then instructs the agent: ``Look at my task list, pick the highest-priority task you can help with, and work on it.''
The agent consults the task board, presents candidate tasks via a multiple-choice interaction, and upon user selection begins executing the chosen task---demonstrating the bidirectional editing loop and full observe--extract--act cycle.
This scenario highlights the system's core novelty: persistent, user-editable organizational knowledge that directly shapes the agent's next action.

\section{Related Work}
\label{sec:related}

\paragraph{Web Agent Systems.}
WebArena~\cite{zhou2024webarena} and OSWorld~\cite{xie2024osworld} established realistic benchmarks, revealing a large gap between LLM-based agents and human operators.
Despite advances in structured reasoning~\cite{agashe2025agents,yang2025agentoccam} and vision-based approaches~\cite{zheng2024seeact,gur2024webagent}, a recent meta-evaluation~\cite{xue2025illusion} found most agents plateau well below human performance; OpAgent~\cite{guo2026opagent} raised the ceiling to 71.6\% but still falls short.
Commercial CUAs~\cite{openai2025cua,anthropic2024computeruse,google2024mariner} demonstrated general-purpose browser automation but do not expose persistent, user-editable organizational context.

\paragraph{Agent Memory Architectures.}
Several works maintain persistent memory for LLM agents: hierarchical memory management~\cite{packer2023memgpt}, code-level skill libraries~\cite{wang2023voyager}, verbal self-reflections~\cite{shinn2023reflexion,zhao2024expel}, and reusable workflow routines~\cite{wang2024workflowmemory}.
These systems keep memory inside the agent, capturing the agent's own experience but not the user's actual work patterns.
\cotomiact{} instead learns from observed browser behavior and externalizes behavioral knowledge as typed, user-editable artifacts; routine user maintenance keeps the agent aligned without explicit synchronization.

\paragraph{Human-AI Collaboration.}
The shared knowledge workspace draws on principles from human-AI collaboration.
Horvitz~\cite{horvitz1999mixed} established foundations for mixed-initiative interaction.
Andrews et al.~\cite{andrews2023sharedmental} showed that shared mental models underpin effective human-AI teaming.
Star and Griesemer~\cite{star1989boundary} introduced \emph{boundary objects}---artifacts shared across different communities yet interpretable by each on its own terms.
\cotomiact{}'s shared artifacts serve as boundary objects: the human sees a task management tool, while the agent sees an action source grounding its next execution. This design achieves transparency by construction---the agent's knowledge is externalized as everyday artifacts rather than requiring post-hoc explanation---and supports the collaborative refinement loop.

\section{Conclusion}
\label{sec:conclusion}

\cotomiact{} demonstrates that a browser-based AI co-worker requires two complementary capabilities: reliable multi-step web execution, and organizational knowledge acquired from everyday user behavior.
The agent scaffold achieves 80.4\% on the WebArena human-evaluation subset---exceeding the reported human baseline---while a progressive abstraction pipeline converts raw browsing activity into behavioral knowledge that both human and agent maintain as shared artifacts through routine use.
A controlled proxy study confirms that task success improves as this behavior-derived knowledge accumulates, validating learning from observation as a practical path to organizational grounding.

\clearpage
\bibliographystyle{ACM-Reference-Format}
\bibliography{references}

\clearpage
\appendix

\section{Scaffold Details}
\label{app:scaffold}

This appendix provides implementation details for key scaffold mechanisms introduced in Section~\ref{sec:architecture}.

\subsection{Test-Time Scaling}

Two mechanisms improve per-task accuracy at the cost of additional inference-time computation.

\paragraph{Best-of-N action selection.}
At each step of the ReAct loop, the scaffold samples $N$ candidate actions from the reasoning model and selects the majority-voted action before execution.
This pre-execution consensus reduces the variance of individual samples and filters out hallucinated or imprecise actions (e.g., clicking the wrong element) without requiring an explicit post-action verifier.
In practice we use moderate $N$ (typically ${\leq}5$) to balance accuracy against latency; adaptive schemes inspired by test-time ensembling~\cite{komiyama2026bestofn} can further improve this trade-off.

\paragraph{Task decomposition.}
The scaffold delegates decomposition to the reasoning model itself: when the model emits a structured plan indicating multiple independent sub-goals within a single user instruction, the scaffold spawns parallel sub-agents, each maintaining a focused context window for its sub-goal~\cite{light2025disc}.
If the model identifies only a single goal, execution proceeds without decomposition.

\subsection{Workspace Retrieval}

In the behavioral-knowledge evaluation (Section~\ref{sec:eval:behavior}), all three knowledge formats---trajectory, script, and insight---are stored as wiki pages in the workspace; the agent retrieves relevant pages via the same \texttt{search\_workspace} mechanism described in Section~\ref{sec:architecture} rather than receiving them in its initial context (see Appendix~\ref{app:protocol} for coverage details).

\section{Behavioral-Knowledge Evaluation Protocol}
\label{app:protocol}

This appendix supplements the controlled study in Section~\ref{sec:eval:behavior} with full protocol details.

\subsection{Benchmark and Task Categories}
We use a six-category subset of WorkArena-L1~\cite{drouin2024workarena} over a ServiceNow instance: \emph{dashboard}, \emph{form}, \emph{knowledge}, \emph{list-filter}, \emph{list-sort}, and \emph{service catalog}.
This split separates list tasks into filtering and sorting categories and excludes navigation/menu tasks, which primarily test UI navigation rather than reusable procedural knowledge.
Each category contains multiple task templates (e.g., \emph{order-sales-laptop}, \emph{order-ipad-mini} within service catalog); evaluation uses 20 random seeds per template.

\subsection{Source Traces}
The ``user behavior'' used as input to the knowledge pipeline consists of successful agent trajectories collected by running a browser agent on WorkArena-L1 tasks.
We treat these trajectories as a proxy for observed user behavior: each trace records the sequence of pages visited, actions taken, and resulting state changes---the same signal the behavior logger would capture from a real user's browsing session.
Real user traces would additionally contain failures, interruptions, and exploratory detours; the agentic ETL pipeline (Section~\ref{sec:architecture}) mitigates such noise by weighting behavioral frequency, so that outlier or erratic actions are unlikely to surface as knowledge artifacts.
Any artifacts that do slip through are subject to the bidirectional curation loop: users inspect wiki pages and task items in the shared workspace and can correct or remove inaccurate entries before the agent acts on them.

\subsection{Domain Coverage}
\emph{Domain coverage} is defined at the category level: $k/6$ coverage means that behavioral knowledge derived from $k$ of the six categories has been written to the shared knowledge workspace as wiki pages.
At each coverage step, one category's worth of knowledge artifacts (approximately 34--50 per category, 277 total at 100\% coverage) is added in its entirety; there is no sub-sampling within categories.
During evaluation, the agent does not receive these artifacts in its initial prompt; instead, it retrieves relevant wiki pages on demand via the \texttt{search\_workspace} tool (Appendix~\ref{app:scaffold}), matching the deployed system's retrieval behavior.

\subsection{Ordering Control}
To control for the effect of which categories are added first, we evaluate six random orderings of the six categories (\texttt{order\_00}--\texttt{order\_05}).
Each data point in Figure~\ref{fig:behavior-coverage} is the mean across these six orderings.

\section{End-to-End Episode: Service Catalog}
\label{app:episode}

Section~\ref{sec:eval:behavior} summarizes the service-catalog episode inline.
Table~\ref{tab:sc-jump} provides the per-ordering breakdown.
In five of six orderings the agent achieves perfect accuracy immediately upon receiving the scripts, regardless of when in the coverage sequence the service-catalog domain is introduced.

For example, the procedural script for \emph{order-sales-laptop} reads:
\begin{quote}\small
\begin{enumerate}[nosep,leftmargin=*]
  \item Navigate to ``Service Catalog.''
  \item Select the ``Hardware'' category.
  \item Select the specific item (e.g., ``Sales Laptop'').
  \item Set the quantity.
  \item For each software option, check the box if required.
  \item Enter additional software requirements.
  \item Click ``Order Now.''
  \item On the confirmation page, copy the request number.
\end{enumerate}
\end{quote}

\begin{table}[H]
  \centering
  \caption{Service-catalog success rate before and after adding catalog-ordering scripts (script format). Each row is one of six random category orderings.}
  \label{tab:sc-jump}
  \small
  \begin{tabular}{lccc}
    \toprule
    & \textbf{Before} & \textbf{After} & \textbf{Final (100\%)} \\
    \midrule
    order\_00 & 65\% & 100\% & 100\% \\
    order\_01 & 75\% & 100\% & 95\% \\
    order\_02 & 70\% & 100\% & 100\% \\
    order\_03 & 80\% & 100\% & 100\% \\
    order\_04 & 90\% & 100\% & 95\% \\
    order\_05 & 70\% & 95\%  & 95\% \\
    \midrule
    \textbf{Mean} & \textbf{75.0\%} & \textbf{99.2\%} & \textbf{97.5\%} \\
    \bottomrule
  \end{tabular}
\end{table}

\end{document}